\documentclass[11pt,a4paper]{article}
\usepackage[hyperref]{emnlp2020}
\usepackage{times}
\usepackage{latexsym}

\usepackage{graphicx}
\usepackage{graphics}
\usepackage{microtype}
\usepackage{booktabs}
\usepackage{amsmath}
\usepackage{enumitem}
\usepackage{multirow}
\usepackage{multicol}
\usepackage{xspace}
\usepackage{amssymb}
\usepackage{listings}
\newcommand{\dataset}{\textsc{KGText}\xspace}

\newcommand{\nop}[1]{}
\newcommand{\model}{KGPT\xspace}

\aclfinalcopy 

\title{KGPT: Knowledge-Grounded Pre-Training \\ for Data-to-Text Generation}

\author{Wenhu Chen, Yu Su, Xifeng Yan, William Yang Wang \\
University of California, Santa Barbara, CA, USA\\
\tt{\{wenhuchen, xyan, william\}@cs.ucsb.edu,} \tt{su.809@osu.edu}\\
}

\date{}

\begin{document}
\maketitle
\begin{abstract}
Data-to-text generation has recently attracted substantial interests due to its wide applications. Existing methods have shown impressive performance on an array of tasks. However, they rely on a significant amount of labeled data for each task, which is costly to acquire and thus limits their application to new tasks and domains. In this paper, we propose to leverage pre-training and transfer learning to address this issue. We propose a knowledge-grounded pre-training (KGPT), which consists of two parts, 1) a general knowledge-grounded generation model to generate knowledge-enriched text. 2) a pre-training paradigm on a massive knowledge-grounded text corpus crawled from the web. The pre-trained model can be fine-tuned on various data-to-text generation tasks to generate task-specific text. We adopt three settings, namely fully-supervised, zero-shot, few-shot to evaluate its effectiveness. Under the fully-supervised setting, our model can achieve remarkable gains over the known baselines. Under zero-shot setting, our model without seeing any examples achieves over 30 ROUGE-L on WebNLG while all other baselines fail. Under the few-shot setting, our model only needs about one-fifteenth as many labeled examples to achieve the same level of performance as baseline models. These experiments consistently prove the strong generalization ability of our proposed framework\footnote{\url{https://github.com/wenhuchen/KGPT}}.   
\end{abstract}

\section{Introduction}
Data-to-text generation, i.e., generating textual description from structured data, is an important task with many real-world applications such as generating weather reports~\cite{liang2009learning}, sports news~\cite{wiseman-etal-2017-challenges}, dialog response~\cite{wen2016multi,dusek2019e2e}, etc. Neural generation models based on different strategies like soft-template~\cite{wiseman2018learning,ye2020variational}, copy-mechanism~\cite{see2017get}, content planning~\cite{reed2018can,moryossef2019step}, and structure awareness~\cite{liu2018table,colin2019generating} have achieved impressive results. However, existing studies are primarily focused on fully supervised setting requiring substantial labeled annotated data for each subtask, which restricts their adoption in real-world applications. 
\begin{figure}[!t]
    \centering
    \includegraphics[width=1.0\linewidth]{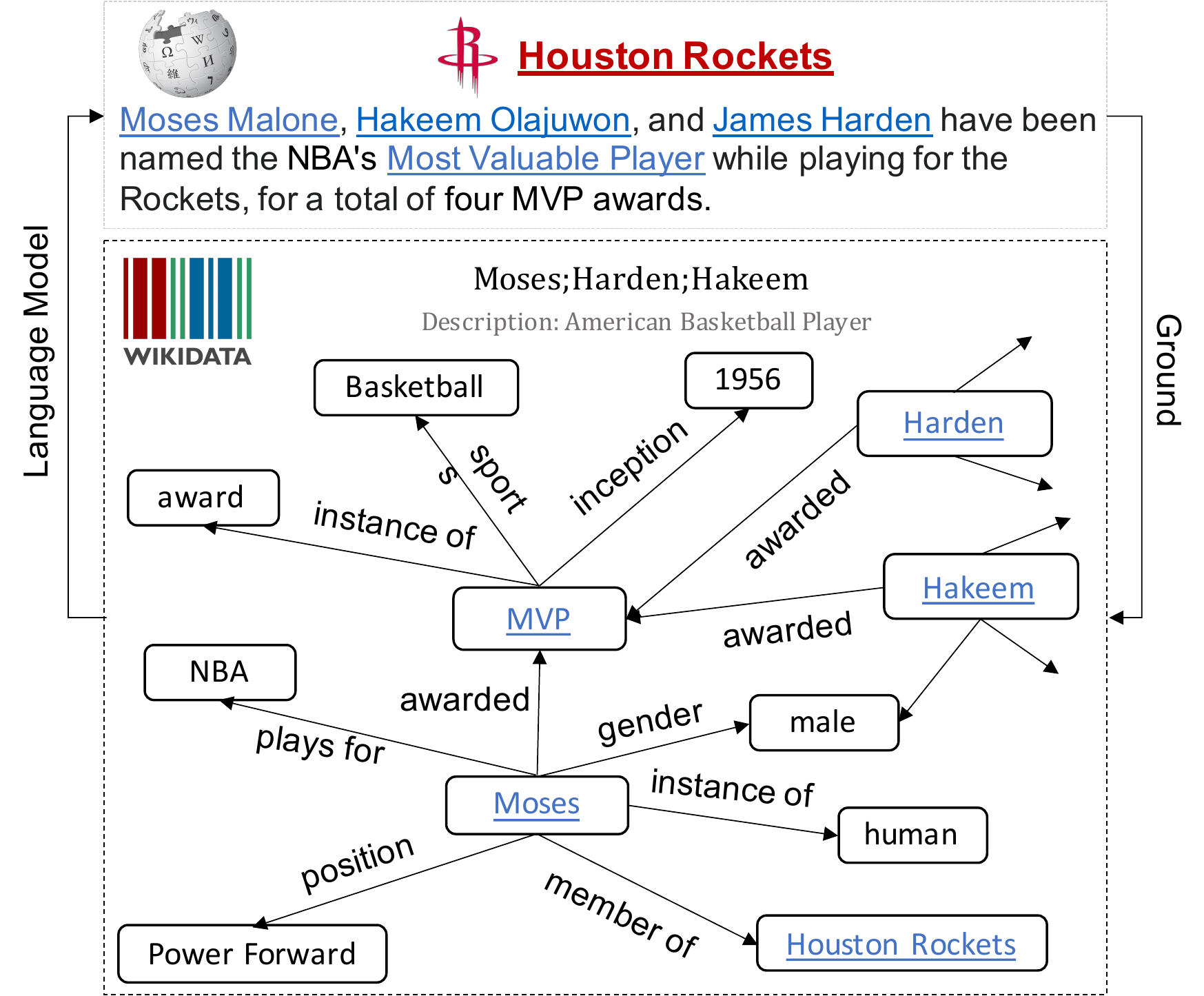}
    \caption{An example from the constructed \dataset, which pairs a hyperlinked sentence from Wikipedia with a knowledge subgraph from WikiData.}
    \label{fig:example}
    \vspace{-2ex}
\end{figure}

In this paper, we are interested in developing a general-purpose model that can easily adapt to different domains/tasks and achieve strong performance with only a small amount or even zero annotated examples. Our model draws inspiration from the recent wave of pre-trained language model~\cite{devlin2019bert,radford2019language,dai2019transformer} to exploit large-scale unlabeled data from the web for pre-training. The data pairs are constructed through the following procedure. We first crawl sentences with hyperlinks from Wikipedia, and then link the hyperlinked entities to WikiData~\cite{vrandevcic2014wikidata} to find their 1-hop knowledge triples. Finally, we build a subgraph based on the linked triples. Such automatic alignment between knowledge graph and texts provides distant supervision~\cite{mintz2009distant} for pre-training but it is bound to be noisy. Therefore, we design a selection strategy and only retain plausible alignments with high semantic overlap. The harvested knowledge-grounded corpus \dataset consists of over 1.8M (knowledge subgraph, text) pairs, as depicted in~\autoref{fig:example}. 

We unify the input of \dataset and down-stream data-to-text tasks into a generalized format and design a novel architecture \model to encode it. We use \dataset to first pre-train \model and then fine-tune it on downstream data-to-text tasks like WebNLG~\cite{shimorina2018handling}, E2ENLG~\cite{dusek2019e2e} and WikiBio~\cite{liu2018table}.
Experimental results demonstrate \model's several advantages: 1) with full down-stream dataset, \model can achieve remarkably better performance than known competitive baselines, 2) with zero training, \model can still achieve a reasonable score on WebNLG. 3) with a few training instances, \model can maintain a high BLEU score while the non-pre-trained baselines only generate gibberish text. A quantitative study shows that our pre-training scheme can reduce annotation costs by roughly 15x to achieve a decent BLEU score of 30.
Our contribution is summarized as follows:\\
    \indent i). We design a distantly supervised learning algorithm to exploit large-scale unlabeled web text to pre-train data-to-text models. \\
    \indent ii).  The proposed pre-training algorithm can bring significant performance under different settings, especially zero-shot and few-shot scenarios.
\section{Related Work}
\paragraph{Data-to-Text Generation}
Data-to-text is a long-standing problem~\cite{kukich1983design,reiter1997building}, which involves generating natural language surface form from structured data. The traditional system is primarily built on a template-based algorithm. Recently, with the development of deep learning, attention has been gradually shifted to end-to-end neural generation models, which achieve significant performances on existing large-scale datasets like WebNLG~\cite{shimorina2018handling}, E2ENLG~\cite{dusek2019e2e}, WikiBio~\cite{lebret2016neural}, ROTOWIRE~\cite{wiseman-etal-2017-challenges}, TOTTO~\cite{parikh2020totto}, LogicNLG~\cite{chen2020logical}, etc. However, these neural generation models are mainly focused on fully supervised learning requiring a huge amount of human annotation for the specific task. Our paper focuses on building a more generalized model architecture, which can adapt to specific tasks well with only a handful of training instances. 

\paragraph{Knowledge-Grounded Language Modeling}
It is of primary importance to ground language models on existing knowledge of various forms. The neural language models~\cite{bengio2003neural} have been shown to well capture the co-occurrences of n-grams in the sentences, but falls short to maintain the faithfulness or consistency to world facts. To combat such an issue, different knowledge-grounded language models~\cite{ahn2016neural,hayashi2019latent,logan2019barack} have been proposed to infuse structured knowledge into the neural language model. These models are mainly focused on enhancing the factualness of unconditional generative models. Inspired by these pioneering studies, we explore the possibility to connect the unconditional generative model with downstream conditional generation tasks. The most straightforward knowledge-intensive conditional generative task is the data-to-text generation, which aims to verbatim given knowledge into lexical format. We demonstrate great potential of the knowledge-grounded pretraining in enhancing the model's factualness on these down-stream data-to-text tasks and believe such language models can be applied to broader range of NLP tasks requiring knowledge understanding. 

\paragraph{Pre-trained Language Model}
Recently, the research community has witnessed the remarkable success of pre-training methods in a wide range of NLP tasks~\cite{devlin2019bert,radford2018improving,radford2019language,dai2019transformer,yang2019xlnet,liu2019roberta,keskar2019ctrl,lan2019albert,lewis2019bart,raffel2019exploring}. These models trained on millions or billions of data unlabeled data demonstrate unprecedented generalization ability to solve related down-stream tasks. However, the existing pre-trained text generation models~\cite{radford2019language,keskar2019ctrl,raffel2019exploring} are initially designed to condition on text input, thus lacking the ability to encode structured inputs. The work closest to our concept is Switch-GPT-2~\cite{chen2019few}, which fits the pre-trained GPT-2 model as the decoder part to perform table-to-text generation. However, their knowledge encoder is still trained from scratch, which compromises the performance. In this paper, we follow the existing paradigm to construct an unlabeled web data for LM pre-training. 

\section{Dataset Construction}
The construction process has two stages, namely the crawling stage and the selection stage:
\subsection{Hyperlinked Sentence Crawling}
We use English Wikidump\footnote{\url{https://dumps.wikimedia.org/}} as our data source. For each Wikipedia page, we split the whole paragraphs into an array of sentences and then tokenize with the nltk toolkit~\cite{loper2002nltk}. We loop through each sentence to keep the sentences with more than 2 Wikipedia anchor links and within the length of 10 and 50. For each candidate sentence, we use its Wikipedia hyperlink to query WikiData~\cite{vrandevcic2014wikidata} and obtain its corresponding entity page\footnote{\url{https://www.wikidata.org}}. We retrieve the neighboring knowledge triples from these entity pages to construct a local 1-hop graph for each entity. The knowledge triples are divided into two types: 1) the object of the triple is also an entity like `(Roma F.C., country, \underline{Italy})', 2) the object of the triple is in plain text like `(Roma F.C., inception, 7 June 1927)'. In the first case, if the object entity also appears in the sentence, we use it as the bridge to build a multi-hop graph like~\autoref{fig:data-selection}. After this step, we collected roughly 4 million pairs in the form of (subgraph, sentence) as the candidate for the following step.
\subsection{Data Selection}
\begin{figure}[!htb]
    \centering
    \includegraphics[width=0.95\linewidth]{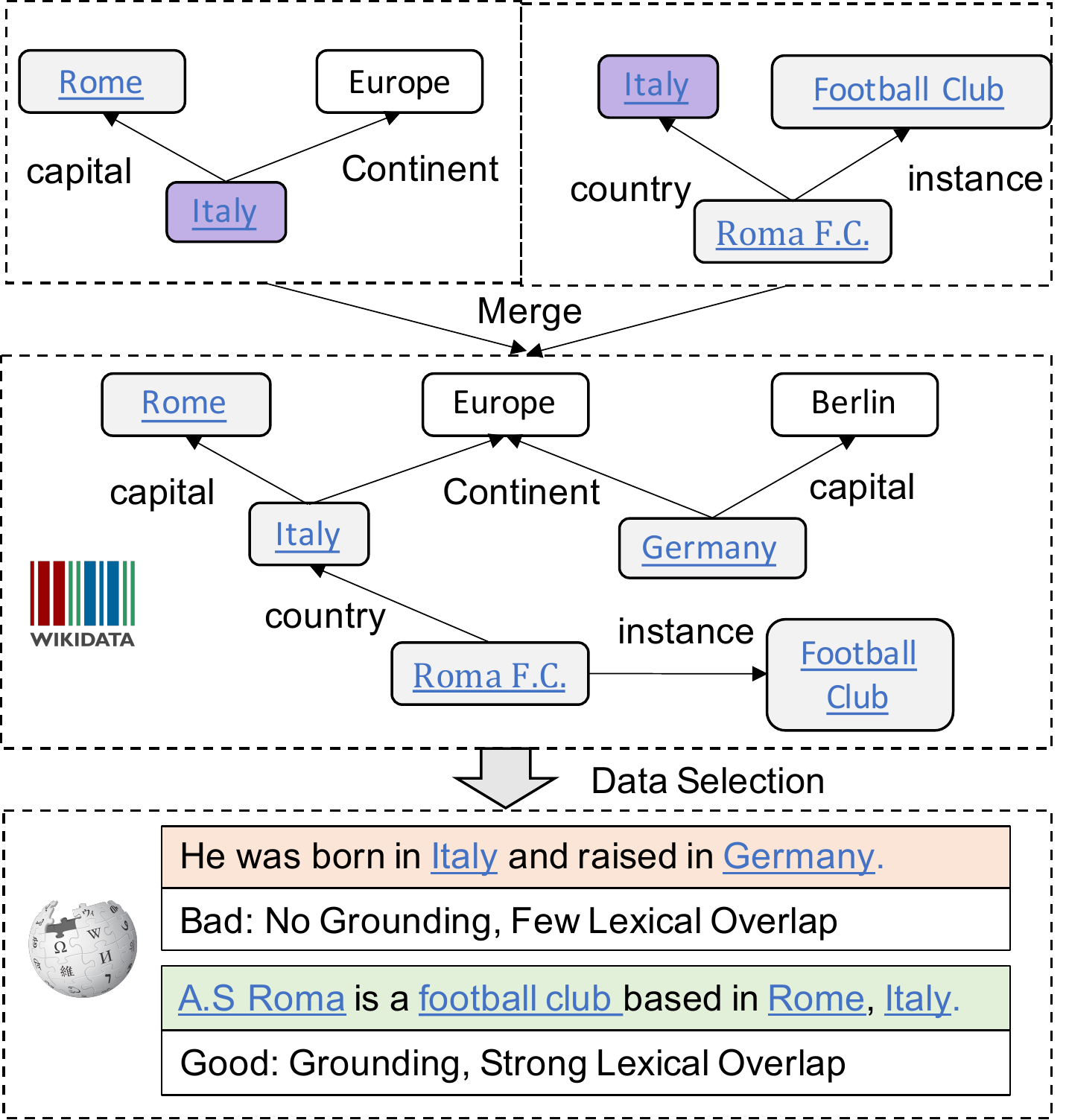}
    \caption{Data denoising procedure for the \dataset.}
    \label{fig:data-selection}
    \vspace{-1ex}
\end{figure}

We observe that the collected pairs are overly noisy with many sentences totally irrelevant to their paired subgraphs. Apparently, these pairs cannot serve our goal to build a knowledge-grounded language model. Therefore, we propose a data selection step to suppress the noise and filter out the data pairs of our interests. An example is depicted in~\autoref{fig:data-selection}, the first sentence does not rely on any information provided by the knowledge graph, while the second sentence has a tight connection to the facts presented in the knowledge graph. Ideally, our proposed strategy should favor the second sentence over the first one.

To achieve this, we propose a simple lexical-based selection strategy to perform data selection. For example, the sentence `He was born ...' in~\autoref{fig:data-selection} has two query words `Italy' and `Germany', we will conduct two rounds of lexical matching. In the first round, we use `Italy' to query its surrounding neighbors in WikiData to the neighboring unigram, i.e. `(Rome, capital, Europe, Continent, Country, Roma F.C)'. We compute the unigram overlap with the original sentence `(He, was, ...)', which is still 0\%. In the second round, we use `Germany' to do the same computation and calculate the lexical overlap, which is still 0\%. So the final averaged grounding score of two rounds is 0\%. We can follow the same procedure to compute the grounding score for the second sentence in~\autoref{fig:data-selection} with four rounds `(AS Rome, FB, Rome, Italy)'. The grounding score is above 30\%, which indicates that the sentence is highly grounded on WikiData subgraph. In this paper, we use a threshold of 0.13, which selects the top 7M `good' sentences from the original 12M Wikipedia corpus.


After the selection step, we obtain a denoised knowledge-grounded corpus \dataset for pre-training. However, there still exist noisy false positives in the corpus, for example, a subgraph contains triple `(Roma F.C., country,  Italy)', which is associated with the text `An Italian player plays for A.S. Roma'. Though the two entities co-occur, they are not meant to describe the fact triple. By applying more strict rules, we can suppress such false positives, but the data capacity could significantly drop consequently. We experimented with different thresholds to balance noise and data capacity and finally decide on a threshold with an acceptable noise degree. The detailed statistics of the \dataset is listed in~\autoref{tab:stat}. 
We held-out 10,000 sentences for both validation and testing to evaluate the pre-trained model.
\begin{table}[!thb]
\centering
\small
\begin{tabular}{lccccc}
\hline
\#Sent & Length & \#Ent & \#Pred & \#Triple & \#Ent/Sent \\
\hline
7M  & 20.2 & 1.8M & 1210 & 16M & 3.0 \\
\hline
\end{tabular}
\caption{Statistics of collected KGText dataset}
\label{tab:stat}
\vspace{-2ex}
\end{table}

\begin{figure*}[!thb]
    \centering
    \includegraphics[width=1.0\linewidth]{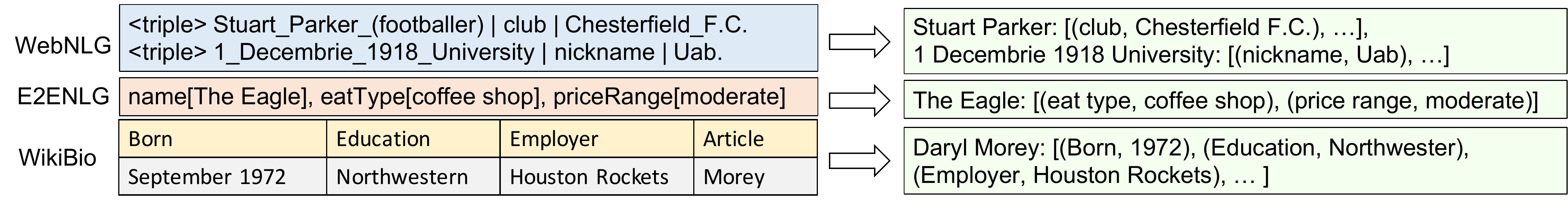}
    \caption{The conversion criterion to unify different structured data input into our generalized format. }
    \label{fig:unification}
    \vspace{-2ex}
\end{figure*}

\section{Model}
We formally define the problem setting and \model's architectures in this section.
\subsection{Problem Setting}
In this paper, we consider inputs from structured data with diverse formats, like knowledge subgraph in \dataset, dialog act in E2E~\cite{dusek2019e2e}, RDF triples in WebNLG~\cite{shimorina2018handling} and tables in WikiBio~\cite{lebret2016neural}. Here we unify them into a generalized dictionary format, which uses keys to represent subjects and values to denote the predicate-object pairs following the subject. We showcase the conversion criteria from structured inputs in different data-to-text datasets into our generalized format in~\autoref{fig:unification}. The generalized input is denoted as $X$, and the output is denoted as $y$. Our model encodes $X$ into a sequence of dense vectors, and then uses the decoder to attend and generate $y$. 

\subsection{Encoder}
\begin{figure*}[!thb]
    \centering
    \includegraphics[width=1.0\linewidth]{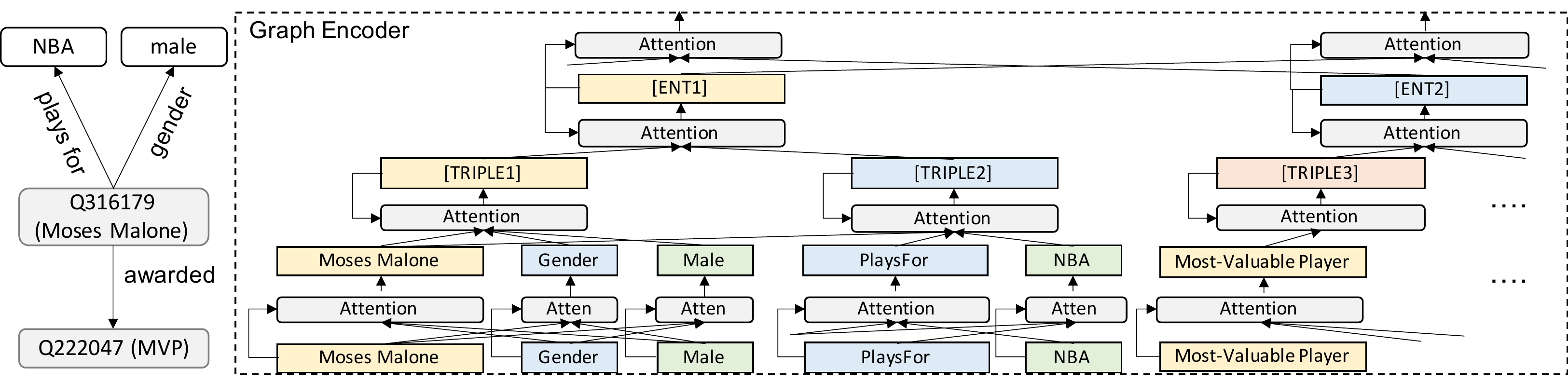}
    \caption{Graph Encoder with hierarchical propagation, where we propagate the information from bottom to top. }
    \label{fig:graph-encoder}
    \vspace{2ex}
    \includegraphics[width=1.0\linewidth]{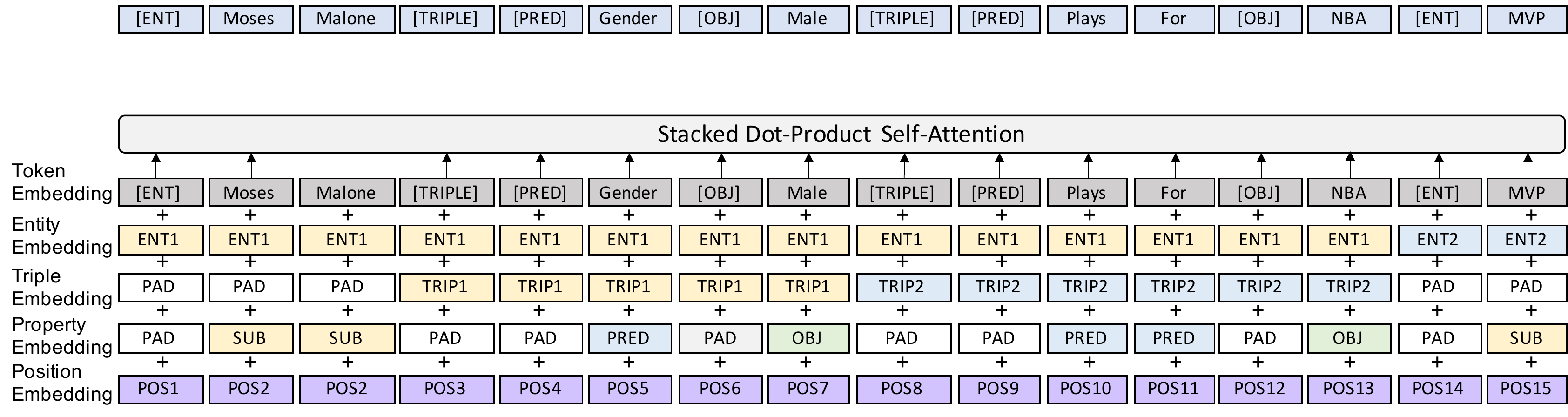}
    \caption{Encoding of the knowledge graph as a sequence using special embedding. }
    \label{fig:sequence-encoder}
\end{figure*}

The encoder network is crucial to our model to capture the highly structured graph input. We mainly experiment with two types of encoders:
\paragraph{Graph Encoder}
This encoder is mainly based on graph attention network~\cite{li2015gated,kipf2016semi,velivckovic2017graph} to explicitly encode the structure information. Specifically, we view each object, predicates, and subjects as the leaf nodes, and add \texttt{[ENT]}, \texttt{[TRIPLE]} as pseudo nodes for message passing purposes. The built graph is depicted in~\autoref{fig:graph-encoder}.

First of all, we initialize the node representation with the averaged embedding of its subword units. For example, the node `Moses Malone' has a representation of (E[Mos] + E[es] + E[Ma] + E[lone]) / 4 with E denoting the embedding. After we obtain the initial node representation, we use message propagation to update the node representations based on neighboring information. 

In the first layer, we exchange the information between nodes inside a triple, e.g., `Moses Malone' receives message from siblings `Gender' and `Male'. In the second layer, we aggregate information from sub/pred/obj nodes to the \texttt{[TRIPLE]} node, e.g., `[TRIPLE1]' receives message from children `Moses, Gender, Male'. In the third layer, we aggregate the information from different \texttt{[TRIPLE]} to the \texttt{[ENT]} node. In the fourth layer, we exchange information between different \texttt{[ENT]} nodes to enhance cross-entity interactions. Formally, we propose to update the representation of the $i$-th node $g_i \in \mathbb{R}^D$ with the multi-head attention network, which aggregates information from neighboring nodes $g_j \in \mathcal{N}_i$ as follows:
\begin{align}
\small
\begin{split}
    \alpha^m_j &= \frac{e^{(W_Q^m g_i)^T (W_K^m g_j)}}{\sum_{j \in \mathcal{N}_i} e^{(W_Q^m g_i)^T  (W_K^m g_j)}} \\
    v &= concat[\sum_{j \in \mathcal{N}_i} \alpha^m_j W_v^m(g_j)] \\
    \hat{g_i} &= LayerNorm(MLP(v + g_i))
\end{split}
\end{align}
where $m$ denotes the $m$-th head in the attention layer, $W^m_Q, W^m_K, W^m_V \in \mathbb{R}^{D \times D}$ are the matrices to output query, key, value vectors for $m$-th head. The attention output $v$ and the residue connection from $g_i$ are fed through the final MLP and LayerNorm to update $i$-th node representation as $\hat{g_i}$. 
The output of graph encoder is denoted as $G \in \mathbb{R}^{n \times D}=\{g_1, \cdots, g_n\}$ with $n$ nodes.

\paragraph{Sequence Encoder}
This encoder is mainly based on transformer~\cite{vaswani2017attention} with special embedding as an auxiliary input to infuse the structure information to the sequence model. The concept of special embedding was initially proposed by BERT~\cite{devlin2019bert}, more recently, it has been adopted by~\citet{herzig2020tapas} to infuse structural information. We visualize the embedding layer in~\autoref{fig:sequence-encoder}, where we leverage additional entity embedding, triple embedding, and property embedding to softly encode the structure of the subgraph as a linearized sequence. For example, the entity embedding can inform the model which entity the current token belongs to, while the triple embedding can indicate which triple the current token belongs to and the property embedding indicates whether the token is a subject, predicate, or a subject. Such an encoding mechanism is designed to softly encode the graph structure into the embedding space for further self-attention. Compared to the graph encoder, the sequence encoder does not enforce the structure as a hard constraint and allows more flexibility for the model to perform cross-triple and cross-entity interactions. Formally, the dot-product self-attention follows the definition of Transformer~\cite{vaswani2017attention}:
\begin{align}
\small
\begin{split}
    &f_{att}(Q, K, V) = softmax(\frac{QK^T}{\sqrt{D}}V)\\
    &G_m = f_{att}(QW_Q^m, KW_K^m, VW_V^m)\\
    &G = MLP(Concat(G_1, \cdots, Gm))
\end{split}
\end{align}
where $Q, K, V$ are the computed from the input embedding, m represents $m$-th head and $f_{att}$ is the core attention function, the final output is denoted as $G \in \mathbb{R}^{n \times D}$ with n denoting the sequence length.  

\subsection{Decoder}
Our decoder architecture is mainly based on Transformer~\cite{vaswani2017attention} and copy mechanism~\cite{see2017get}. At each decoding time step, the model has a copy gate $p_{gen}$ to select $y_i$ should be generated from the vocabulary $w \in \mathcal{V}$ or copied from the input tokens $x$:
\begin{align}
\small
\begin{split}
    \alpha_j = \frac{e^{o_i^T G_j}}{\sum_{j'} e^{o_{i}^T G_{j'}}}, \quad p_{gen} = \sigma(MLP(o_i)) \\
    P(y_i=w) = p_{gen} P_{voc}(w) + (1 - p_{gen}) \sum_{j:x_j = w} \alpha_j
\end{split}    
\end{align}
where $o_i$ is the last layer hidden state of the decoder at $i$-th time step, $\alpha_j$ is the copy probability over the whole input token sequences $x$. 

\subsection{Optimization}
As we have defined our encoder-decoder model, we will simply represent it as $p_{encdec}(x)$ to output a distribution over word $y_i \in \mathcal{V}$ at the $i$-th time step. During pre-training, we optimize the log-likelihood function on $D_{KGText}$. After pre-training, we convert the downstream task's input into the defined dictionary format and denote the dataset as $D_{down}$, and then further optimize the log-likelihood objective with $\theta$ initialized from the pre-training stage. 

\begin{figure}[!thb]
    \centering
    \includegraphics[width=0.9\linewidth]{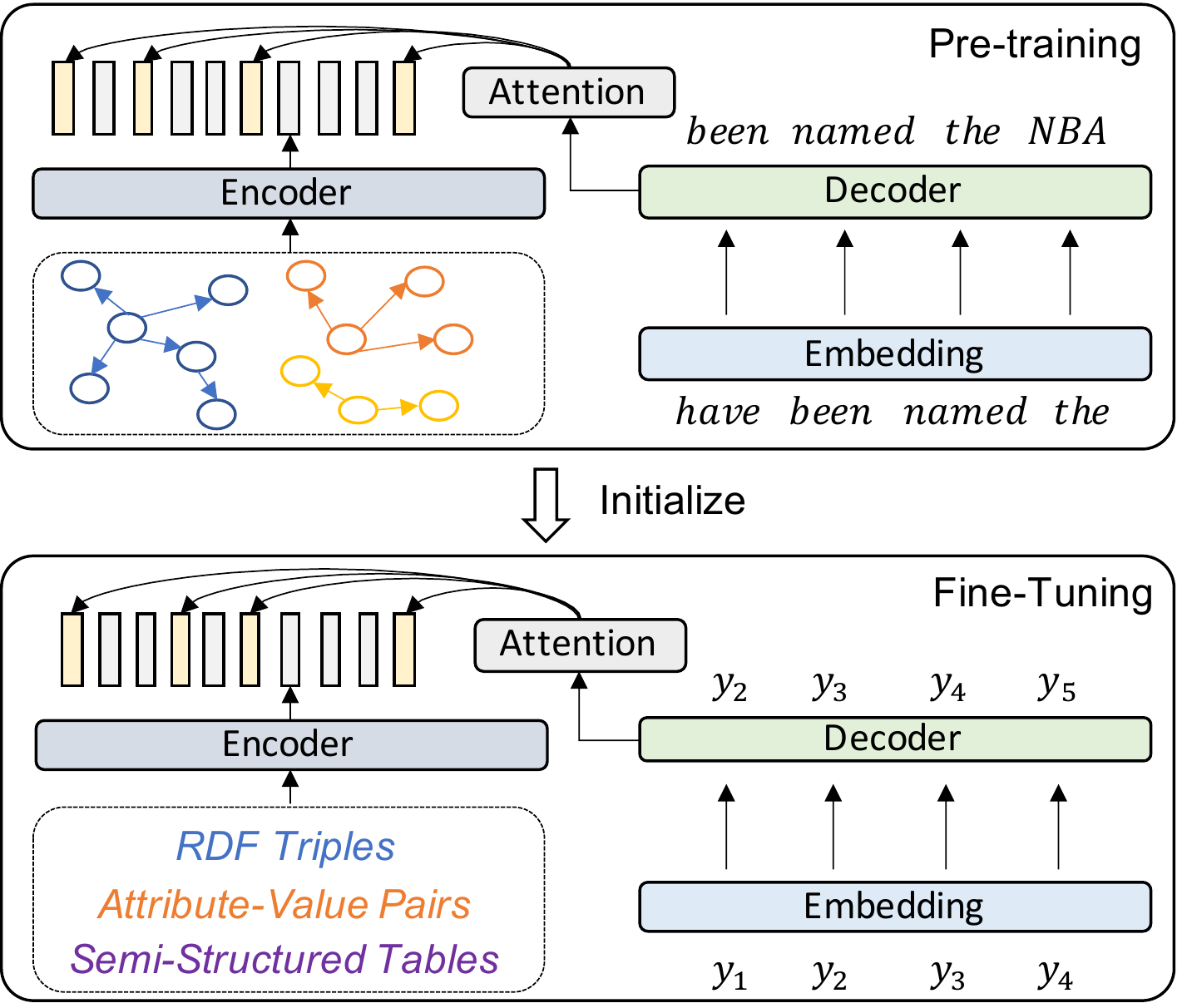}
    \caption{Overall pre-training and fine-tuning procedures for \model. The downstream knowledge data formats are converted into the generalized format. }
    \label{fig:pre-trained-model}
\end{figure}

The pre-train and fine-tuning procedure is displayed in~\autoref{fig:pre-trained-model}, where we first use \dataset to pre-train \model, and then fine-tune with different types of inputs using the standard auto-regressive log-likelihood objective. 


\section{Experiments}
We experiment with three different down-stream tasks, which covers various table-to-text applications to verify the generalization  capability of \model. Besides the fully supervised learning, we also evaluate zero-shot and few-shot learning.

\subsection{Datasets}
We use WebNLG~\cite{shimorina2018handling}, E2ENLG~\cite{dusek2019e2e} and WikiBio~\cite{lebret2016neural} to evaluate the performance of \model. Their basic statistics are listed in~\autoref{tab:datasets}. WebNLG and E2ENLG are both crowd-sourced by human annotator while WikiBio is from the Web.
\begin{table}[!thb]
\small
\begin{tabular}{lllll}
\hline
Dataset & Train & Val & Test & Input \\
\hline
WebNLG  & 34,338 & 4,313 & 4,222 & RDF Triple \\
E2ENLG & 42,061  & 4,672 & 4,693 & Dialog Act \\
WikiBio & 582,657 & 72,831 & 72,831 & Table \\
\hline
\end{tabular}
\caption{Statistics of different data-to-text datasets}
\label{tab:datasets}
\vspace{-2ex}
\end{table}
\paragraph{WebNLG}
This dataset~\cite{shimorina2018handling} aims to convert RDF triples into a human annotated textual description. We use the recent release 2.0 from GitLab\footnote{\url{https://gitlab.com/shimorina/webnlg-dataset}}. It contains sets with up to 7 triples each along with one or more references. The number of KB relations modeled in this scenario is potentially large and generation involves solving various subtasks (e.g. lexicalisation and aggregation). As the input RDF triples were modified from the original triples in DBPedia, we first need to check whether there are seen triples in pre-training dataset \dataset. We verify that there is zero RDF triple seen during pre-training though 31\% entities are seen. Therefore, we can confirm the comparison with other baselines is still fair given no information from test/dev is leaked.  
\paragraph{E2ENLG}
This dataset~\cite{dusek2019e2e} aims to convert dialog act-based meaning representation into a spoken dialog response. It aims to provide higher-quality training data for end-to-end language generation systems to learn to produce more naturally sounding utterances. In this dataset, each meaning representation is associated with on average with 8.65 different reference utterances. 
\paragraph{WikiBio}
This dataset~\cite{lebret2016neural} aims to generate the first sentence of biography description based on a Wikipedia infoboxes table, with each table associated with only one reference. Unlike the previous two human-annotated datasets from different domains, WikiBio is also scraped from Wikipedia. Therefore, we filtered out the instances of \dataset from the first paragraph of the biography domain to ensure no overlap or leakage about Wikibio's dev/test set. 

\subsection{Experimental Setup}
We apply the standard GPT-2~\cite{radford2019language} tokenizer from Hugginface Github\footnote{\url{https://github.com/huggingface/transformers}} to tokenize the text input, which has a vocabulary of over 50K subword units. We test with both graph encoder and sequence encoder. We set their hidden size to 768 and stack 6 layers for both encoder and decoder with 8 attention heads. During pre-training, we run the model on \dataset on 8 Titan RTX GPUs with a batch size of 512 for 15 epochs using Adam~\cite{kingma2014adam} optimizer with a learning rate of 1e-4. The pre-training procedure takes roughly 8 days to finish. We use a held-out validation set to select the best checkpoint. During fine-tuning, we use a learning rate of 2e-5.

In our following experiments, we compare with the known best models from different datasets. As none of these models are pre-trained, we also add Template-GPT-2~\cite{chen2020logical} and Switch-GPT-2~\cite{chen2019few} as our pre-trained baselines. Both models apply GPT-2~\cite{radford2019language} as the generator to decode description from a table. For the ablation purposes, we list the performance of all non-pre-trained \model to see the performance gain brought by pre-training alone. All the best models are selected based on the validation set score, and the numbers are reported in the following tables are for test split. For evaluation, we report the performance with BLEU~\cite{papineni2002bleu}, METEOR~\cite{banerjee2005meteor} and ROUGE-L~\cite{lin-2004-rouge} using e2e-metric\footnote{\url{https://github.com/tuetschek/e2e-metrics}}. It's worth noting that we perform comprehensive data contamination studies in the following experiments to make sure the pre-training data contains very little overlap with the test split in downstream tasks. We filter out potentially information-leaking pages during the data crawling process. 

\nop{
\begin{table}[!thb]
\small
\begin{tabular}{lccccc}
\hline
Model & Enc & \#Enc & \#Dec & Head & Dim \\
\hline
\model-Graph & Graph & 5  & 5  & 8  &  512\\
\model-Seq & Seq  & 5  & 5  & 8  &  512\\
\hline
\end{tabular}
\caption{Implementation details of different \model.}
\label{tab:model-gallery}
\end{table}
}

\subsection{Preliminary Study on \dataset}
In the preliminary study, we evaluate our pre-trained model's performance on the held-out set of \dataset to conduct ablation study over \model. Specifically, we investigate 1) which encoding mechanism is better, 2) whether we need copy mechanism or copy supervision. As demonstrated in~\autoref{tab:ablation}, we observe that the trivial difference between two encoder designs. With the copy mechanism, \model can greatly decrease the perplexity. However, supervising the copy attention does not have much influence on the performance. Therefore, in the following experiments, we will run experiments for both encoding schemes with a copy mechanism without copy loss. 
\begin{table}[!thb]
\centering
\small
\begin{tabular}{lcc}
\hline
Model & BLEU-4 & Perplexity \\
\hline
\model-Graph  & 24.71  &  4.86 \\
\model-Graph + Copy Loss  & 24.77  &  4.91 \\
\model-Graph w/o Copy  & 22.69  &  7.23 \\
\hline
\model-Seq    & 24.49  &  4.95 \\
\model-Seq + Copy Loss   & 24.31  &  4.93 \\
\model-Seq w/o Copy  & 22.92  & 7.11  \\
\hline
\end{tabular}
\caption{Ablation Study on held-out set of \dataset.}
\label{tab:ablation}
\end{table}

\subsection{Fully-Supervised Results}
We experiment with \model under the standard fully-supervised setting to compare its performance with other state-of-the-art algorithms.
\paragraph{WebNLG Challenge}
We list WebNLG's experimental results in~\autoref{tab:web}, here we compare with the known models under the unconstrained setting. The baseline models~\cite{shimorina2018handling} uses sequence-to-sequence attention model~\cite{luong2015effective} as the backbone and propose delexicalization and copy mechanism to enhance model's capability to handle rare items from the input. The GCN model~\cite{marcheggiani2018deep} uses graph convolutional neural encoder to encode the structured data input. Its implementation is from Github\footnote{\url{https://github.com/diegma/graph-2-text}}. As can be seen, \model without pre-training already achieves better performance than the GCN baseline. With pre-training, the performance is further boosted by 1-2 BLEU-4, which reflects the effectiveness of our method. 
\begin{table}[!thb]
\small
\begin{tabular}{lccc}
\hline
Model                & BLEU  & METEOR & ROUGE \\
\hline
Seq2Seq$^{\dagger}$           & 54.0  & 37.0    & 64.0 \\
Seq2Seq+Delex$^{\dagger}$   & 56.0  & 39.0    & 67.0 \\
Seq2Seq+Copy$^{\dagger}$    & 61.0  & 42.0    & 71.0 \\
GCN  & 60.80  & 42.76   & 71.13 \\
\hline
\model-Graph w/o Pre   & 62.30 & 44.33  & 73.00 \\
\model-Seq w/o Pre   & 61.79 & 44.39  & 72.97 \\
\hline
\model-Graph w/ Pre    & 63.84    &   46.10    &  74.04 \\
\model-Seq w/ Pre    &  \textbf{64.11}     &   \textbf{46.30}    &  \textbf{74.57} \\
\hline
\end{tabular}
\caption{Experimental results on WebNLG's test set, w/ Pre refers to the model with pre-training, otherwise it refers to the model training from scratch. ${\dagger}$ results are copied from~\citet{shimorina2018handling}. }
\label{tab:web}
\vspace{-3ex}
\end{table}

\paragraph{E2E Challenge}
We list E2ENLG's experimental results in~\autoref{tab:e2e}, here we compare with the state-of-the-art systems on the leaderboard of E2E challenge\footnote{\url{http://www.macs.hw.ac.uk/InteractionLab/E2E/}}. These baselines methods are based on neural template model~\cite{wiseman2018learning}, syntax-enhanced algorithms~\cite{duvsek2016sequence}, slot alignment~\cite{juraska2018deep} and controlling mechanism~\cite{elder2018e2e}. As is seen from the table, \model can beat the SOTA systems by a remarkable margin. Overall, the improvement brought by pre-training is roughly 0.5-1.0 in terms of BLEU-4, which is less significant than WebNLG. Such a phenomena is understandable given that this dataset contains limited patterns and vocabulary in the input meaning representation, a full training set over 40K instances is more than enough for the generation model to memorize. In the following few-shot experiments, we will show the strength of \model to generate high-quality faithful descriptions with only 0.1\% of training data.
\begin{table}[!t]
\small
\begin{tabular}{lccc}
\hline
Model              & BLEU  & METEOR & ROUGE \\
\hline
NTemp              & 55.17 & 38.75 &  65.01 \\
TGen               & 65.93 & 44.83  & 68.50   \\
SLUG2SLUG          & 66.19 & 44.54  & 67.72   \\
Adapt              & 67.37 & 45.23  & 70.89   \\
\hline
\model-Graph w/o Pre & 66.47 & 44.20  & 67.78   \\
\model-Seq w/o Pre & 67.67  & 45.33 & 70.39 \\
\hline
\model-Graph w/ Pre    &  67.87     & 44.50   &  70.00    \\
\model-Seq w/ Pre     &  \textbf{68.05}      & \textbf{45.80}    & \textbf{70.92} \\
\hline
\end{tabular}
\caption{Experimental results on E2E's test set. NTemp is from~\citet{wiseman2018learning}, TGen is from~\citet{duvsek2016sequence}, SLUG2SLUG is from~\citet{juraska2018deep} and Adapt is from~\citet{elder2018e2e}. }
\label{tab:e2e}
\vspace{-2ex}
\end{table}
\paragraph{WikiBio Dataset}
We list WikiBio's experimental results in~\autoref{tab:wikibio} and compare with models like Table2Seq\cite{bao2018table}, Order Planning~\cite{sha2018order}, Field Gating~\cite{liu2018table}, Background-KB Attention~\cite{chen2019enhancing}, Hybrid Hierarchical Model~\cite{liu2019hierarchical} trained with multiple auxiliary loss functions. We also train Template-GPT-2 on this dataset to observe pre-trained model's performance. As can be seen from the table, \model can achieve better results than the mentioned baseline models. Pre-training can yield an improvement of roughly 0.5 BLEU-4. As this dataset trainin/testing have similar table schema and the large number of training instances already teach the model to memorize the generation patterns, exploiting an external corpus of on par size (1.8M) does not bring a significant boost. So is the template-GPT-2~\cite{chen2020logical}, which performs on par with Field Gating~\cite{liu2018table}. However, in the few-shot setting, we will show the 25+ BLEU gain brought by pre-training.
\begin{table}[!thb]
\centering
\small
\begin{tabular}{lc}
\hline
Model              & BLEU \\
\hline
Table NLM~\cite{lebret2016neural}               & 34.70 \\
Table2Seq~\cite{bao2018table}          & 40.26 \\
Order Planning~\cite{sha2018order}              & 43.91 \\
Field-Gating~\cite{liu2018table}       & 44.71 \\
KBAtt~\cite{chen2019enhancing}              & 44.59 \\
Hierarchical+Auxiliary Loss~\cite{liu2019hierarchical}  & 45.01 \\
\hline
Template-GPT-2 & 44.67 \\
\model-Graph w/o Pre & 44.64  \\
\model-Seq w/o Pre &  44.58 \\
\hline
\model-Graph w/ Pre    & \textbf{45.10} \\
\model-Seq w/ Pre     & 45.06 \\
\hline
\end{tabular}
\caption{Experimental results on WikiBio's test set. }
\label{tab:wikibio}
\vspace{-2ex}
\end{table}

\subsection{Few-Shot Results}
The few-shot learning setting aims to study the potential of the proposed pre-training to decrease annotation labor in data-to-text generation tasks. Under this setting, we not only compare with non-pre-trained baselines to observe how pre-training can benefit the model's few-shot learning capability but also compare with other pre-trained LM~\cite{chen2019few,chen2020logical} to see the benefit of \model over existing pre-trained LM. 
\begin{table}[!thb]
\centering
\small
\begin{tabular}{lcccc}
\hline
Model & 0.5\% & 1\% & 5\% & 10\%\\
\hline
Seq2Seq & 1.0 & 2.4  & 5.2  & 12.8  \\
Seq2Seq+Delex  & 4.6 & 7.6  & 15.8  & 23.1  \\
\model-Graph w/o Pre & 0.6 & 2.1 & 5.9 & 14.4 \\
\model-Seq w/o Pre  & 0.2 & 1.7 & 5.1 & 13.7\\
Template-GPT-2  & 8.5   & 12.1 & 35.3  & 41.6 \\
\hline
\model-Graph w/ Pre & \textbf{22.3} & \textbf{25.6} & \textbf{41.2} & \textbf{47.9} \\
\model-Seq w/ Pre   & 21.1  & 24.7  & 40.2 &  46.5\\
\hline
\end{tabular}
\caption{Few-shot results on WebNLG's test set. }
\label{tab:webnlg-few-shot}
\vspace{2ex}
\begin{tabular}{lcccc}
\hline
Model & 0.1\% & 0.5\% & 1\% & 5\%\\
\hline
TGen  & 3.6  &  27.9  &  35.2   & 57.3\\
\model-Graph w/o Pre & 2.5  & 26.8 & 34.1  & 57.8\\
\model-Seq w/o Pre  &  3.5 & 27.3   & 33.3 & 57.6\\
Template-GPT-2  & 22.5   & 47.8 & 53.3  & 59.9 \\
\hline
\model-Graph w/ Pre & 39.8  & \textbf{53.3} & \textbf{55.1}  & \textbf{61.5} \\
\model-Seq w/ Pre  & \textbf{40.2}  & 53.0 & 54.1  &  61.1 \\
\hline
\end{tabular}
\caption{Few-shot results on E2ENLG's's test set. }
\label{tab:e2enlg-few-shot}
\vspace{-2ex}
\end{table}
\paragraph{WebNLG \& E2ENLG Dataset}
In these two datasets, we use 0.1\%, 0.5\%, 1\%, 5\%, 10\% of training instances to train the model and observe its performance curve in terms of BLEU-4.

For WebNLG challenge, the few-shot situation will pose a lot of unseen entities during test time. From ~\autoref{tab:webnlg-few-shot}, we can observe that the delexicalization mechanism can remarkably help with the few-shot situation. However, the improvement brought by delexicalization is much weaker than our proposed pre-training. Under the 5\% setting, while the non-pre-trained baselines are only able to generate gibberish text, pre-trained \model can maintain a high BLEU score over 40.0 due to its strong generalization ability.  

For E2E challenge, the task is comparatively simpler with rather limited items. From~\autoref{tab:e2enlg-few-shot}, we can observe that TGen~\cite{duvsek2016sequence} is achieving similar performance as our non-pre-trained \model, they both perform quite well even under 1\% training instances. However, after we further reduce the training samples to roughly 0.1\%, the baseline models fail while pre-trained \model still maintains a decent BLEU over 40.0. 

\paragraph{WikiBio Dataset}
In this dataset, we adopt the same setting as Switch-GPT-2~\cite{chen2019few} and Pivot~\cite{ma2019key} to use 50, 100, 200 and 500 samples from the training set to train the generation model. From the results in~\autoref{tab:wikibio-few-shot}, we observe that \model can achieve best scores and outperform both Template-GPT-2 and Switch-GPT-2 under most cases. Though Template-GPT-2 is getting slightly better score with 500 training samples, the overall performance on three datasets are remarkably lower than \model, especially under more extreme cases. It demonstrates the advantage of our knowledge-grounded pre-training objective over the naive LM pre-training objective. 
\begin{table}[!thb]
\centering
\small
\begin{tabular}{lcccc}
\hline
Model & 50 & 100 & 200 & 500 \\
\hline
Field-Infusing   &  1.3  & 2.6     &  3.1  &  8.2 \\
\model-Graph w/o Pre  & 0.2  & 1.1  & 3.8  & 9.7 \\
\model-Seq w/o Pre   & 0.6  & 1.7  & 3.0  & 8.9 \\
\hline
Pivot$^{\dagger}$   &    7.0  &   10.2   &   16.8   &    20.3    \\
Switch-GPT-2$^{\dagger}$  &  17.2 &  23.8    &   25.4       &     28.6  \\
Template-GPT-2  & 19.6   & 25.2 & 28.8 & \textbf{30.8} \\
\hline
\model-Graph w/ Pre & \textbf{24.5}  & 27.5  &  28.9  & 30.1  \\
\model-Seq w/ Pre   & 24.2  &  \textbf{27.6}  &  \textbf{29.1} &  30.0 \\
\hline
\end{tabular}
\caption{Few-shot results on Wikibio's test set. ${\dagger}$ results are copied from~\citet{chen2019few}. }
\label{tab:wikibio-few-shot}
\vspace{-2ex}
\end{table}
\paragraph{Quantitative Study}
We further investigate how much sample complexity \model can reduce. Specifically, we specify a BLEU-4 score and vary the training data size to observe how much training samples are required to attain the performance. We specify BLEU=30 as our standard and display our results in~\autoref{tab:sample-efficiency}.
\begin{table}[!thb]
\centering
\small
\begin{tabular}{lcccc}
\hline
Model & WebNLG & E2ENLG & WikiBio\\
\hline
\model w/o Pre  & $\sim$10000  &  $\sim$300    & $\sim$8000   &  \\
\model w/ Pre  & $\sim$700   &   $\sim$20    &  $\sim$500  &  \\
Ratio &   14x & 15x  & 16x \\
\hline
\end{tabular}
\caption{Required number of training samples to reach designated BLEU on different dataset.}
\label{tab:sample-efficiency}
\vspace{-1ex}
\end{table}
We compute the ratio of sample quantity to characterize the benefits from pre-training. Roughly speaking, pre-training can decrease the sample complexity for training by 15x, which suggests the great reduction rate the annotation cost with pre-trained \model to achieve the desired `promising' performance. 

\subsection{Zero-Shot Results}
We further evaluate \model's generalization capability under the extreme zero-shot setting and display our results for WebNLG in~\autoref{tab:webnlg-zero-shot}. As can be seen, all the non-pre-trained baselines and Template-GPT-2 fail under this setting, while \model can still manage to generate reasonable outputs and achieve a ROUGE-L score over 30. Given that no input knowledge triples in WebNLG were seen during pre-training, these results reflect \model's strong generalization ability to cope with out-of-domain unseen knowledge inputs.
\begin{table}[!thb]
\centering
\small
\begin{tabular}{lccc}
\hline
Model & BLEU & METEOR & ROUGE\\
\hline
All Baselines & 0 & 0 & 1.2 \\
Template-GPT-2 & 0.3  & 0.5 & 3.4 \\
\hline
\model-Graph w/ Pre   & 13.66 & 19.17 & 30.22 \\
\model-Seq w/ Pre & 13.86 & 20.15 & 30.23 \\
\hline
\end{tabular}
\caption{Zero-shot results on WebNLG's test set. }
\label{tab:webnlg-zero-shot}
\end{table}

\begin{figure}[!thb]
    \centering
    \includegraphics[width=1.0\linewidth]{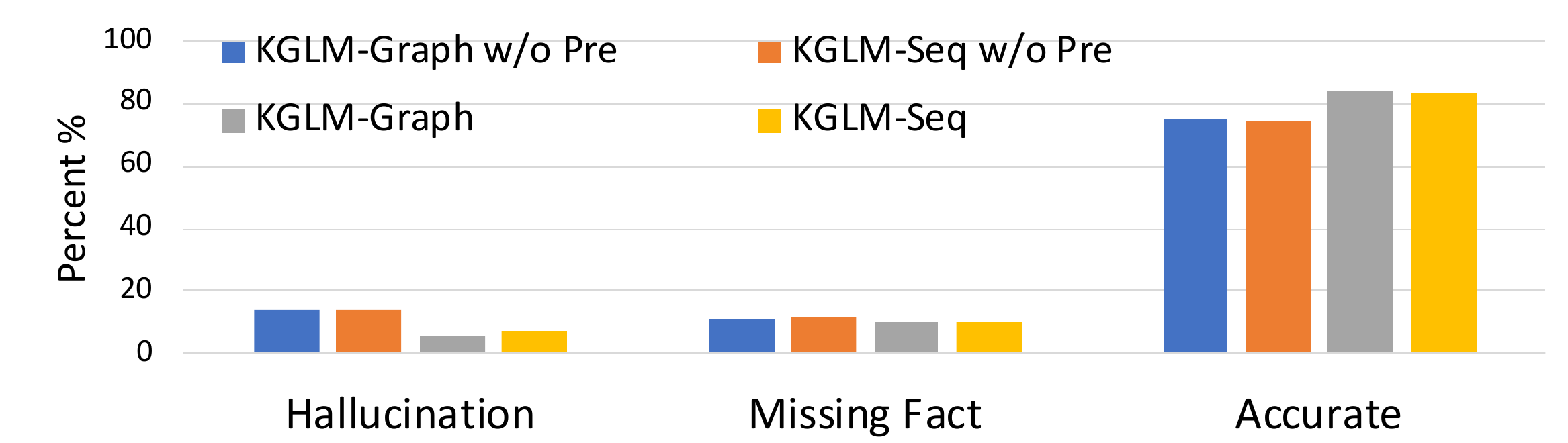}
    \caption{Human evaluation of the factual consistency of different models on WebNLG samples.}
    \label{fig:human-evaluation}
    \vspace{-1ex}
\end{figure}
\subsection{Human Evaluation}
We conduct human evaluation to assess the factual accuracy of the generated sentences. Specifically, we sample 100 test samples from WebNLG and observe the model's factual consistency with given fact triples.  We use AMT to distribute each generated sentence to four high-quality workers (95\% approval rate, 500+ approved jobs) to choose from the three ratings. The majority voted rating is the final rating. 
We compare four different systems, i.e., non-pre-trained and pre-trained \model. Conditioned on the fact triples, we categorize the generated samples into the following categories: 1) hallucinating non-existing facts, 2) missing given facts without hallucination, 3) accurate description of given facts. We visualize the results in~\autoref{fig:human-evaluation}, from which we observe that pre-trained \model are less prone to the known hallucination issue and generate more accurate text. The human evaluation suggests that pre-training can enhance the model's understanding over rare entities, thus reducing the over-generation of non-existent facts.

\subsection{Conclusion}
In this paper, we propose a pre-training recipe to exploit external unlabeled data for data-to-text generation tasks. Our proposed model has achieved significant performance under zero-shot and few-shot settings. Such a framework provides a plausible solution to greatly reduce human annotation costs in future NLG applications.

\section*{Acknowledgement}
The authors would like to thank the anonymous reviewers for their thoughtful comments. This research is sponsored in part by NSF IIS 1528175, we also want to thank their financial support.

\bibliography{emnlp2020}
\bibliographystyle{acl_natbib}
\clearpage
\appendix

\section{Learning Curve}
\begin{figure}[!thb]
    \centering
    \includegraphics[width=1.0\linewidth]{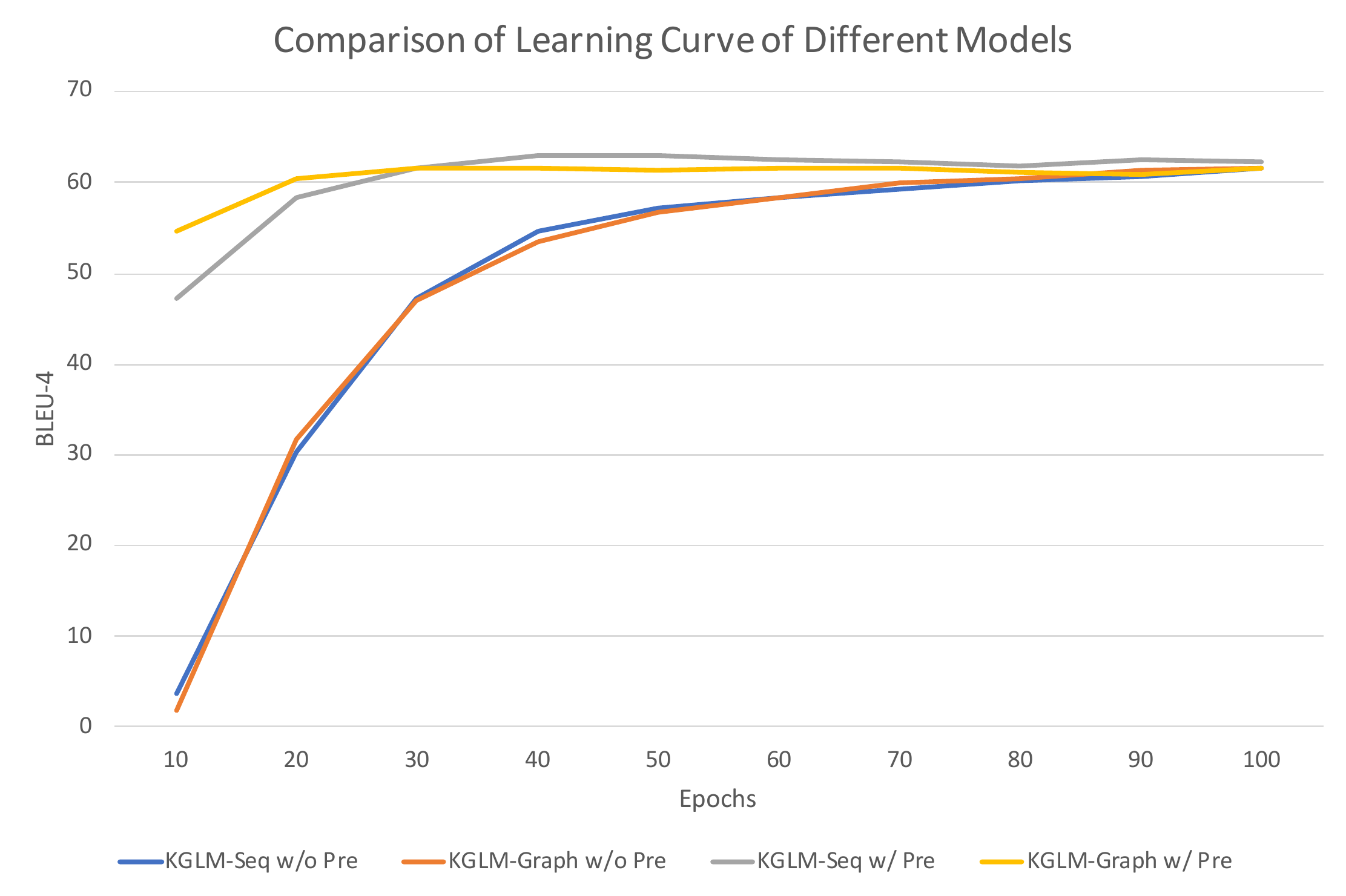}
    \caption{The learning curve of different models during training for the WebNLG dataset. }
    \label{fig:curve}
\end{figure}

Here we observe the learning trend of both non-pre-trained and pre-trained models by evaluating the validation BLEU at each epoch end, here we show our findings in~\autoref{fig:curve}. As can be seen from the figure, the pre-trained model converges much faster to the best score. More specifically, it only takes 20 epochs for the model to reach BLEU-4 over 60 while it takes 80-90 epochs for a non-pre-trained model to reach equivalent performance.

\begin{figure*}[!tbh]
    \centering
    \includegraphics[width=1.0\linewidth]{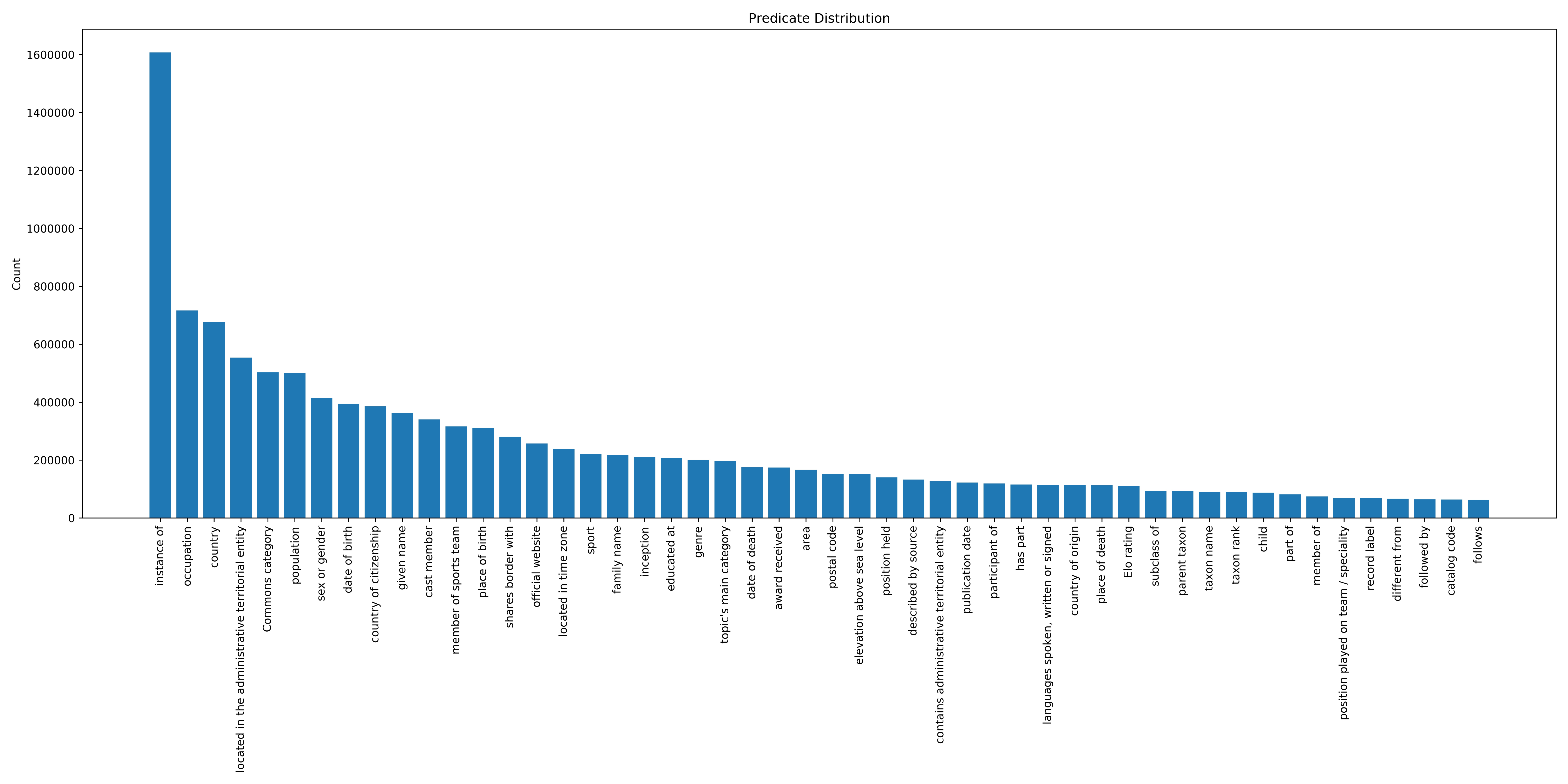}
    \caption{Predicate distribution over the knowledge triples in \dataset. }
    \label{fig:pred-distribution}
\end{figure*}
\section{Predicate Distribution}
Here we demonstrate the most popular predicates in~\autoref{fig:pred-distribution}. As can be seen, the most popular predicates are `instance of', `occupation', `country', `located in', etc. There are over 1000 predicates in our dataset, which covers the commonly seen categories in different domains like politics, athletics, music, news, etc.

\section{Case Study}
\begin{figure*}[!tbh]
    \centering
    \includegraphics[width=1.0\linewidth]{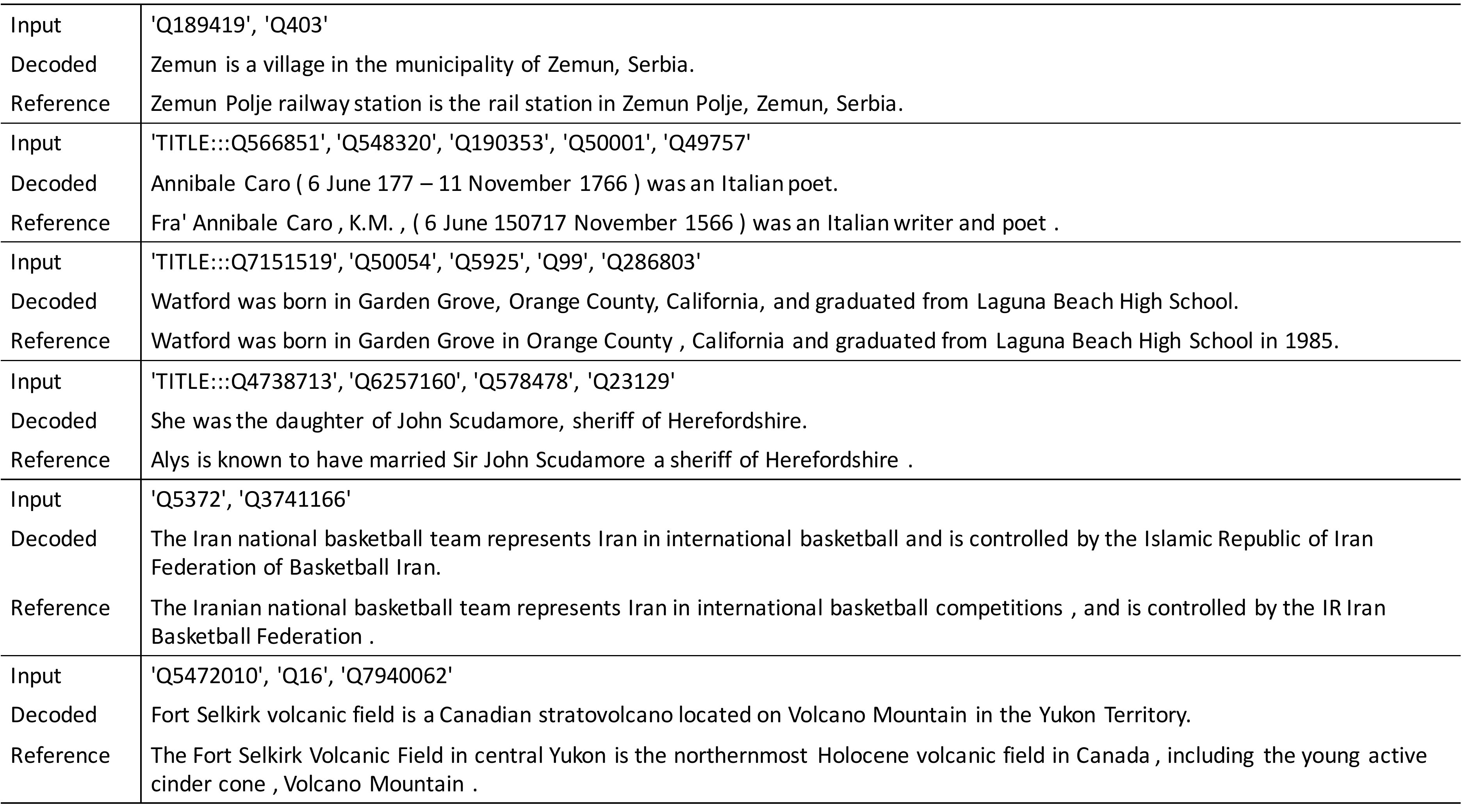}
    \caption{Randomly generated samples from \dataset, where the inputs are the WikiData entities, you can search it online to see it information. For example, the entity 'Q403' and its fact triples can be seen from \url{https://www.wikidata.org/wiki/Q403}. }
    \label{fig:examples}
\end{figure*}

Here we demonstrate some empirical study over the generated samples from our models in~\autoref{fig:examples}. As can be seen, \model has developed a really strong generation capability to output fluent and coherent sentences. In the first line, the decoded sentence is mostly correct, just the name of `municipality' should be `Belgrade' rather than `Zemun' itself according to \url{https://www.wikidata.org/wiki/'Q189419}. In the second line, the sentence is mostly correct, the error comes from the end date of Annibale. The third sentence is completely correct. The fourth sentence also suffers from a factual error, the relationship should be `married' rather than `daughter'. 

From these sentences, it's understandable that the model can achieve reasonable zero-shot performance on the WebNLG dataset given that WebNLG also comes from a similar domain. The case study reveals that our generation model though generates fluent and relevant sentences from the given knowledge triples, the groundedness is still questionable with quite an amount of hallucination issues. 

\end{document}